# Local Interpretable Model Agnostic Shap Explanations for machine learning models

P. Sai Ram Aditya[1,2], Mayukha Pal[1,$]

[1]ABB Ability Innovation Center, Asea Brown Boveri Company, Hyderabad 500084, India.
[2]Department of Artificial Intelligence, Indian Institute of Technology Hyderabad, Kandi, Sangareddy, Telangana 502285, India.

[$]Corresponding author:

Dr. Mayukha Pal
R&D Principal Program Manager
ABB Ability Innovation Center
10th Floor, Western Aqua, Kondapur
Hyderabad – 500084, TS, India.
Tele: +91-9866161632
Email: Mayukha.pal@in.abb.com

## Abstract

With the advancement of technology for artificial intelligence (AI) based solutions and analytics compute engines, machine learning (ML) models are getting more complex day by day. Most of these models are generally used as a black box without user interpretability. Such complex ML models make it more difficult for people to understand or trust their predictions. There are variety of frameworks using explainable AI (XAI) methods to demonstrate explainability and interpretability of ML models to make their predictions more trustworthy. In this manuscript, we propose a methodology that we define as Local Interpretable Model Agnostic Shap Explanations (LIMASE). This proposed ML explanation technique uses Shapley values under the LIME paradigm to achieve the following (a) explain prediction of any model by using a locally faithful and interpretable decision tree model on which the Tree Explainer is used to calculate the shapley values and give visually interpretable explanations. (b) provide visually interpretable global explanations by plotting local explanations of several data points. (c) demonstrate solution for the submodular optimization problem. (d) also bring insight into regional interpretation e) faster computation compared to use of kernel explainer.

## Significance

Shapley value is a concept from game theory which has many desirable properties useful in XAI methods but its computation for most of the complex ML models is slow with accuracy issues. Tree explainer algorithm introduced by Lundberg et. al. computes exact shapley values efficiently and accurately on tree-based models. Our proposed method leverages this algorithm using a decision tree and calculate shapley values to interpret any model hence observed to be much faster as it is locally faithful following the LIME paradigm. The method also helps in global interpretation following the SP-LIME paradigm and additionally gives insight into regional interpretation.

## 1. Introduction

Nowadays, we find in almost all domains Machine Learning is used. Due to the increase in computation power, cloud technologies and its availability to more people, domains every day, ML methods are being used extensively not only by individuals or industries but also in significant domains like defense, healthcare, finance etc [1-2]. These domain users have little freedom for error from ML models hence it is worth a discussion how much of a black box ML model could be used for such decision making [3]. In view of this growing discussion for the ML models, XAI techniques tries to bring many solutions to these questions [4-5]. Machine learning research in the area of XAI is evolving rapidly [6-8].

There are many XAI methods evolved in past and continual research discussion in this area recently for better model development to make it more explainable, interpretable and universal [9-11]. Considerable research is being done to evaluate various XAI methods [12-14]. Local interpretable model agnostic explanations (LIME) is a popular XAI method proposed by Marco et al. [15]. It trains an easily interpretable model $g$ to be locally faithful to the predictions of a complex model $f$ where g is further interpreted for local explanations. As $g$ has the knowledge of the decision boundary around the point it was trained in an example of a classification problem thus interpreting $g$ is same as interpreting $f$ locally. $g$ was taken as a simple logistic regressor in this work with its weights giving the explanations. This method is simple yet very effective.

Shapley values is another XAI method with its roots from the game theory. It was introduced by Lloyd Shapley in 1951 for which he later got a Nobel prize in year 2012. Shapley value is known to have many desirable properties that many other methods lack hence it's the most used XAI method [16-18] but calculating them has been a problem in the field of AI. There were no methods to calculate shapley values on complex models accurately and efficiently for ML Models until 2020 when Lundberg et al. proposed the Tree Explainer algorithm [19] that computes shapley value exactly on any Tree based model efficiently.

In this paper, we propose a methodology to make use of LIME approach while leveraging the goodness of shapley value simultaneously. A simple decision tree is trained to be locally faithful for any complex model $f$. The tree explainer algorithm is used to give shapley explanation values for any local point. With use of shapley and LIME, our method is also model-agnostic hence it works without any knowledge about the model. Model-agnostic methods generally need a black box machine learning model which could predict from a desired input. This is not only convenient for the user but also helps in protecting the confidentiality of those complex model $f$ when needed. When an important decision to be made in domains like health, finance or defense, individual interpretation is important to trust the decision of models. Our proposed method compliments such scenario by calculating the contributions of different features that makes the decision while demonstrating explainability. The proposed method helps in visualizing [20] these contributions and hence could help establish trust of end user for the ML model prediction, decision.

Many a times individual interpretation is not enough when a model needs to be field deployed. It is important that we have a global understanding of the model. The proposed method also helps in visualizing the global explanations by plotting local explanations of several points. Local interpretation of all the samples gives a global inference of the model but it may not be efficient repetitively in all scenarios. We also demonstrate our method usability in implementing Submodular Pick (SP). SP algorithm helps in selecting samples as diverse as possible based on the budget and thus efficiently perform global interpretation using those samples. The manuscript is organized with section 2 describing details of the materials used in our proposed methodology while section 3 details the results and discussions using LIMASE. The section 4 of the manuscript discusses about our inference and future scope.

## 2. Methods and Materials:

Let $f$ be the complex machine learning model that needs to be interpreted and $g$ be the simple interpretable model we chose, to mimic $f$ locally following the LIME paradigm. In our proposed methodology $g$ is a decision tree that we used.

## 2.1 Choice of using Shapley Values

Shapley value concept from game theory has been very reliable, desirable in XAI methods ever since it was recently proposed [16, 21-22]. Some of its desirable properties are listed below:

Here V denotes the value function which when given a feature returns its importance. S is the set of features while N is the total number of features. $\emptyset(V, i)$ denotes the Shapley value of $i$ th feature when value function V is used. In our case V is the ML predictor $f$. We compute shapley values as:

$$\emptyset(V, i) = \sum_{s \subseteq S/\{i\}} \frac{|s|! * (N-|s|-1)!}{N!} * (V(s \cup i) - V(s)) \quad (1)$$

1) Efficiency: The sum of the contribution i.e the shapley value of each feature must add up to the model's prediction subtracting the average prediction of the model i.e the base value in SHAP.

$$\text{Axiom 1: } \sum_{i \in N} \emptyset(V, i) = V(N) - V(\{\}) \quad (2)$$

2) Consistency: If a feature A is consistently contributing more to the prediction than feature B, then shapley values faithfully reflects it hence the shapley value of A will be higher than that of B.

$$\text{Axiom 2: } V1(S \cup i) - V1(S) \geq V2(S \cup i) - V2(S) \; \forall S$$
$$\Rightarrow \emptyset(V1, i) \geq \emptyset(V2, i) \quad (3)$$

3) Missingness: A feature that does not contribute to the prediction must be attributed a zero value.

$$\text{Axiom 3: } V(S \cup i) = V(S) \forall S$$
$$\Rightarrow \emptyset(V, i) = 0 \quad (4)$$

4) Symmetry: Symmetric features contribute equivalently to the final prediction and hence have the same contribution values.

All these desirable properties of Shapley values makes it one of the most important methods suitable for explainable AI. Calculating exact shapley values has been difficult. Many methods were proposed to calculate approximate shapley values. Recently, it was demonstrated Tree Explainer could calculate exact shapley values efficiently. To leverage these awesome features, our architecture uses shapley in LIME paradigm with decision tree as the underlying tree explainer model.

## 2.2 Choice of using Decision Tree

In LIME, *g* was either a logistic regressor in case of classification or a linear regressor in case of regression which were interpretable by themselves. In our proposed method, we chose decision tree as g to use the tree explainer. Decision trees are known to perform well both in regression and classification tasks. They are robust and less sensitive to outliers. They are not only capable of giving a well fitted model but also works well with the tree explainer.

In LIME the Fidelity-Interpretability trade-off was discussed where fidelity is the measure of how faithful *g* is to *f* locally and interpretability is a measure of how easy it is to interpret *g*. In our proposed method, the interpretability of *g* comes from the fact that it is a decision tree, and we use tree explainer to calculate the exact shapley values efficiently. Fidelity comes from how well *g* fits *f* locally. Decision trees are known to work well with less data which was observed in our implementation as well. Decision trees mimic the actual network very well locally. The trade-off is discussed more in section 2.3.1.

## 2.3 Local Interpretation

| **Algorithm 1** | Shapley Explanation Value computation for Proposed Method |
|---|---|
| **Inputs:** | Model f, Number of samples N, sample x, weight definition w. |

$P \leftarrow \{\}$
$for\ i \in \{1,2,3,\ldots .N\}$:
　　$p_i \leftarrow$ sample_around(x)
　　$P \leftarrow P \cup \{p_i, f(p_i), w(p_i)\}$
End for
Train a decision tree g on P.
Shapley Values $\leftarrow$ Tree Explainer(g, x)
Return Shapley Values

**Algorithm 1 Explanation**

Let x be the point where we wish to interpret f. N is the number of perturbations around x that we use to train our decision tree g. Then we sample N points around x with x' be one of them and obtain f(x') for all these points to build a dataset $D = \{(x_i', f(x_i'), w(x_i'))\}$. Here w is the weight function that calculate weights of the samples in D based on their distance from x. Now train g on D. Calculate shapley values on g at x using SHAP's Tree Explainer. These values give an inference about how different features contribute to predict f(x) for x. SHAP Force plot helps towards visual interpretation of these values. The above method is repeated for multiple x's and plotted a SHAP summary plot for all the values by providing a global interpretation on model f. The steps are:

1. Select the sample *x* where we wish to explore local interpretation.
2. Sample N points around *x* using any distribution and get $Z_x$.
3. Construct array of weights
4. Then, calculate *f(z)* for every *z* in $Z_x$ and get $Z_y$.
5. Now train a Decision Tree (*g*) with $Z = (Z_x, Z_y)$ as train data and *W* as sample weights.
6. Use Tree Explainer on *g* to calculate shapley values at sample *x*.
7. Draw force plot using shapley values and interpret the contributions of various features for predicting *f(x)*.
8. Repeat above steps for multiple points and draw the summary plot for all the values for the global interpretation.

### 2.3.1 Kernel Width($\sigma$) and Local Fidelity-Interpretability tradeoff:

A popular choice of w is to reduce the weights exponentially with 1 at x and 0 at infinite distance from x which we practice in our experimentation too. w could be expressed as $\exp(-d(x, x')^2 / \sigma^2)$ where d is the distance function and σ is the kernel width. The Kernel width decides how the weights are distributed across different perturbations based on their distance from *x*. For a small kernel width, the weights are more concentrated around *x* and as we increase the kernel width, these weights are slowly diluted, increasing the weights of farther points. Higher the kernel width, closer the base value of *g* to *f*. When the kernel width is infinite, training g on D is equivalent to mimic f globally. For our shapley values of *g* to be closer to that of which would be calculated on *f*, kernel width needs to be high. But increasing the kernel width makes *g* lose its local faithfulness. Hence, there is a tradeoff between interpretability and local fidelity. It was observed that taking the kernel width five times the standard deviation of the data across different dimensions, gave the best result.

**Analysis:** Though the interpretability and fidelity tradeoff is a problem it could be still used to our advantage in different use cases. By varying the kernel width as required, we could find out how the prediction of *x* is different from its local neighborhood by restricting the training of *g* to a smaller area around x. For example, when predicting the GDP of different cities across the country based on some features, using shapley values directly on *f* shows the features of a city *x* that drives its prediction with respect to the whole country's average GDP. When we wish to know the features of the city that is driving its predictions with respect to the state's average GDP, we could achieve it by varying the kernel width suitably again.

## 2.4 Submodular Pick:

With explanation of local interpretation in the previous sub section, here we explain the submodular optimization problem that could be solved using our proposed method. Submodular pick helps in global interpretation of the model by selecting individual instances that are diverse in their dependence on the features hence gives non redundant information. Generally, one would not require interpreting all the instances for a global understanding of the model.

| **Algorithm 2** | Submodular pick implementation in proposed algorithm |
|---|---|
| **Inputs:** | Samples X, Budget G |
| | for all $x_i \in X$ do |
| |      $S_i \leftarrow$ Shap_explain($x_i$) |
| | end for |
| | for $j \in \{1....d\}$ do          #d is the number of features of our instances |
| |      $I_j \leftarrow \sqrt{\sum_{i=1}^{n} \lvert S_{ij} \rvert}$ |
| | End for |
| | SP ← { } |
| | While $\lvert SP \rvert < G$ do |
| |      $SP \leftarrow SP \cup argmax_i \, q(SP \cup \{i\}, S, I)$ |
| | End while |
| | Return SP |

**Algorithm 2 explanation**

The algorithm takes instances X, predictor f and budget G as inputs where X is taken to be a random subset of the available data. G denotes the number of individual instances that we wish the algorithm to return. First, we compute the shapley explanation values for all the instances of X and then calculate the importance value $I_j$ for all features as shown in Algorithm 2. Then similar approach like LIME is used to select G instances from the function q as shown in Algorithm 2.

$$q(SP, S, I) = \sum_{j=1}^{d} 1_{[\exists i \in SP: S_{ij} > 0]} I_j \quad (5)$$

SHAP summary plots give better visualization and a global idea of the model when drawn.

## 3. Results and Discussion

To demonstrate efficacy of the proposed method, we have considered public available dataset for our model and discussed its results. We have discussed in this section the importance of various plots before showcasing results from our proposed method for the ease of understanding.

## 3.1 SHAP Plots

We make use of the following two plots to visualize the results from our implementation.

- Force Plot: SHAP force plot shows the features that pushes the prediction of an input away or towards the base value of the predictor and also further quantify it. This plot helps in understanding the contribution of different features in predicting a certain value for a single input. Bigger arrow for a feature indicates bigger contribution of the feature in pushing the prediction in the direction of that arrow. Refer figure 1.
- Summary plot: SHAP summary plot has the shapley values on the x-axis with the features on the y-axis while the color of a point gives its feature value. This plot helps us to understand how shapley values change with the values of different features. Refer figure 2.

## 3.2 Dataset

We have considered two datasets a classification and a regression data to demonstrate efficacy of our proposed method in achieving its objective.

**Classification data**

For classification, we make use of the public available data of mobile price in our analysis [23]. This is a multi-class classification data with label 0 for low cost, label 1 for medium cost, label 2 for high cost and label 3 for very high cost with features of the mobile are ram size, battery power, height and width, etc.

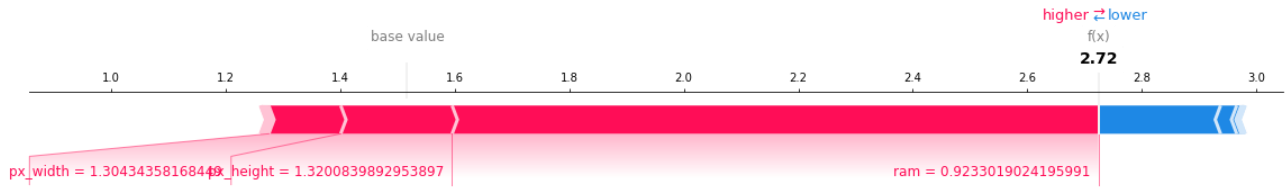

Figure 1: SHAP Force Plot for local interpretation using Tree Explainer. From the figure, we observe the feature ram size contributed the most in forcing the prediction to the positive direction followed by the height feature.

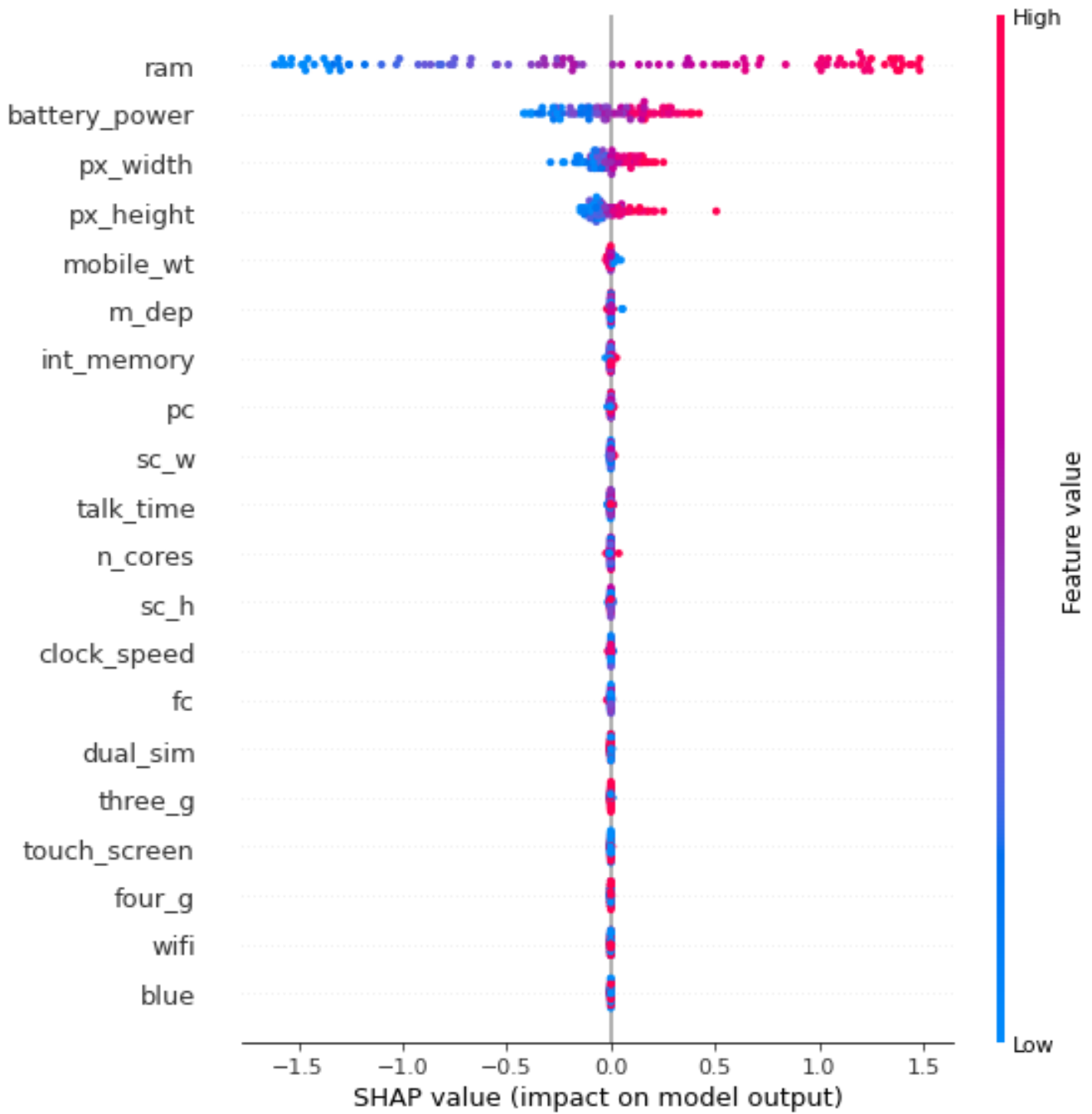

Figure 2: SHAP summary plot for 100 randomly selected training samples using Tree Explainer demonstrating shapley value contribution on feature importance.

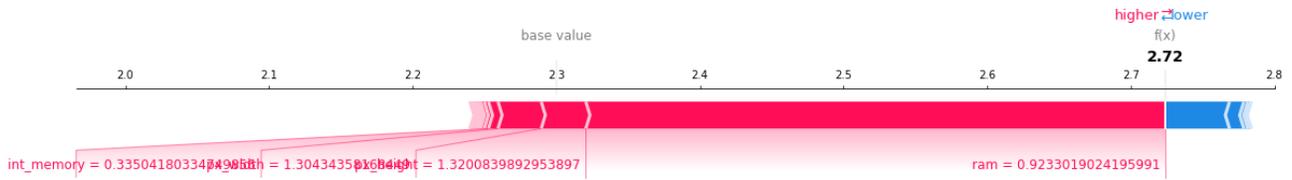

Figure 3: SHAP Force Plot for local interpretation using proposed method LIMASE showing results as in case of the tree explainer method.

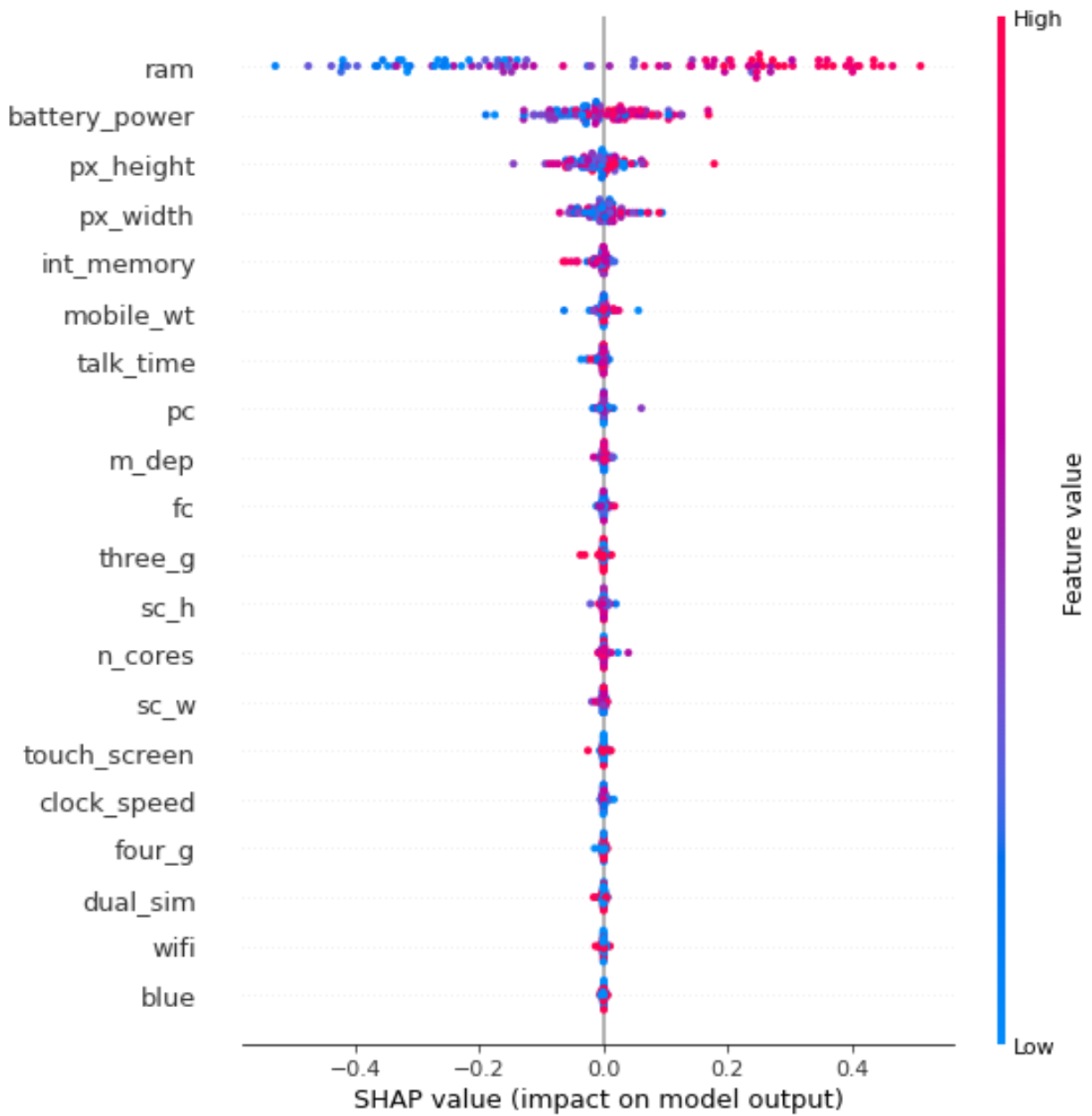

Figure 4: SHAP summary plot for 100 training samples using proposed method.

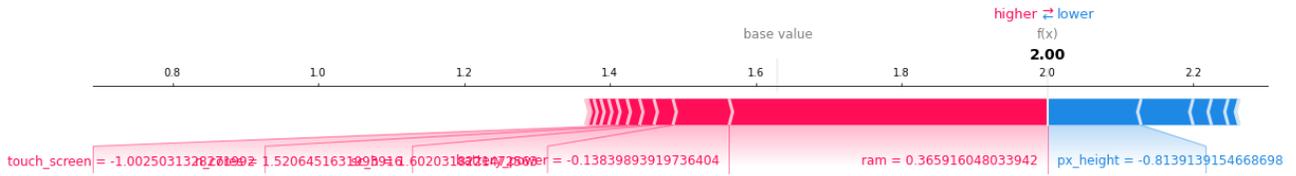

Figure 5: SHAP Force Plot for local interpretation using kernel explainer on a MLP Classifier. It shows the feature ram contributed the most in forcing the prediction to the positive direction followed by the feature battery power.

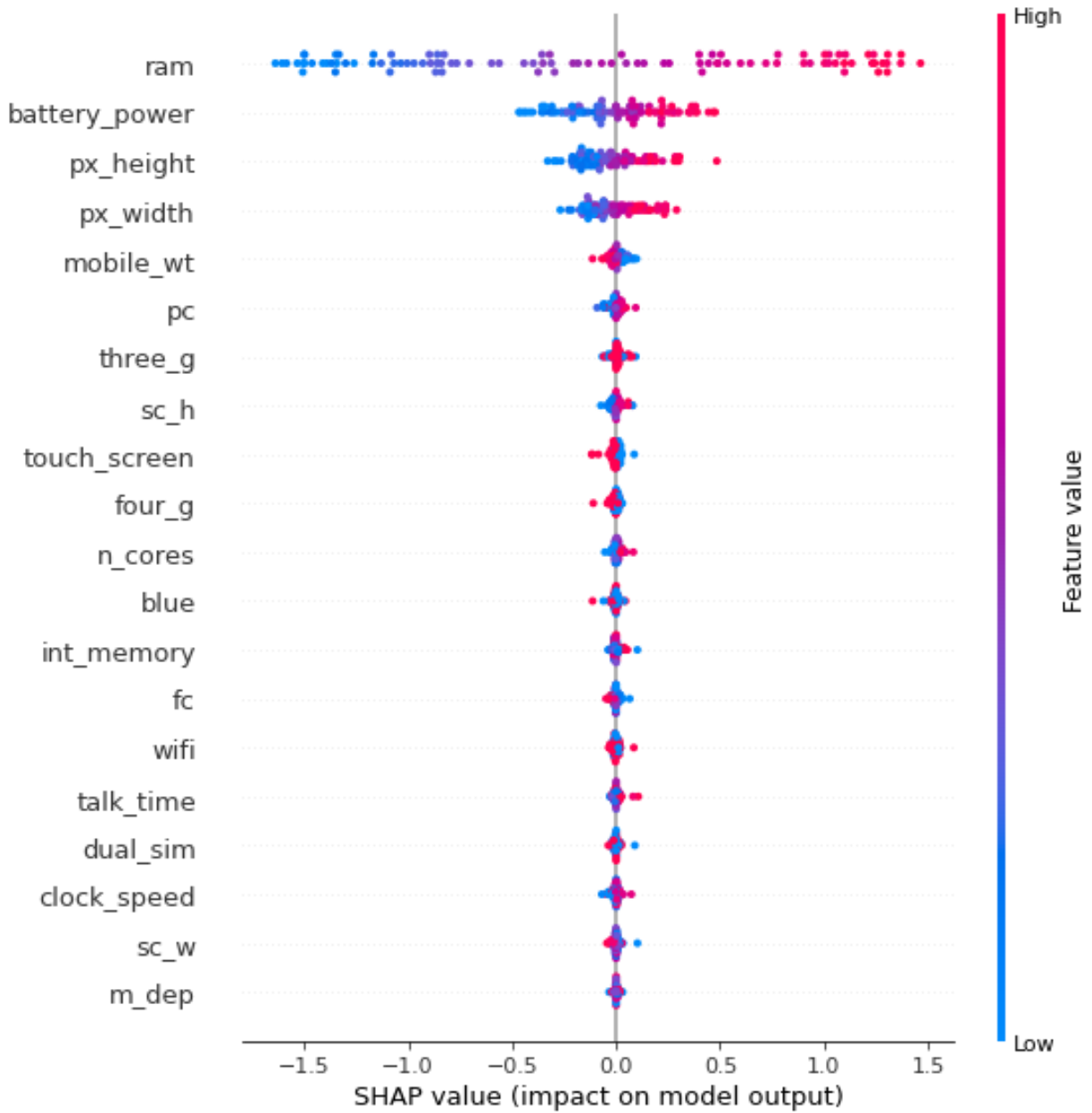

Figure 6: SHAP summary plot for 100 training samples using tree explainer on the MLP Classifier.

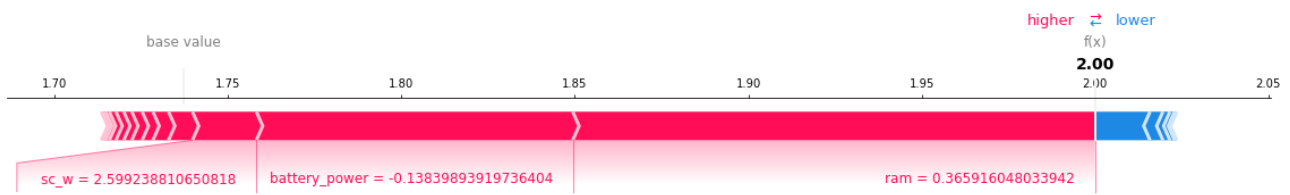

Figure 7: SHAP Force Plot for local interpretation using our proposed method on MLP Classifier. It shows the feature ram size contributed the most in forcing the prediction in the negative direction followed by the feature battery power.

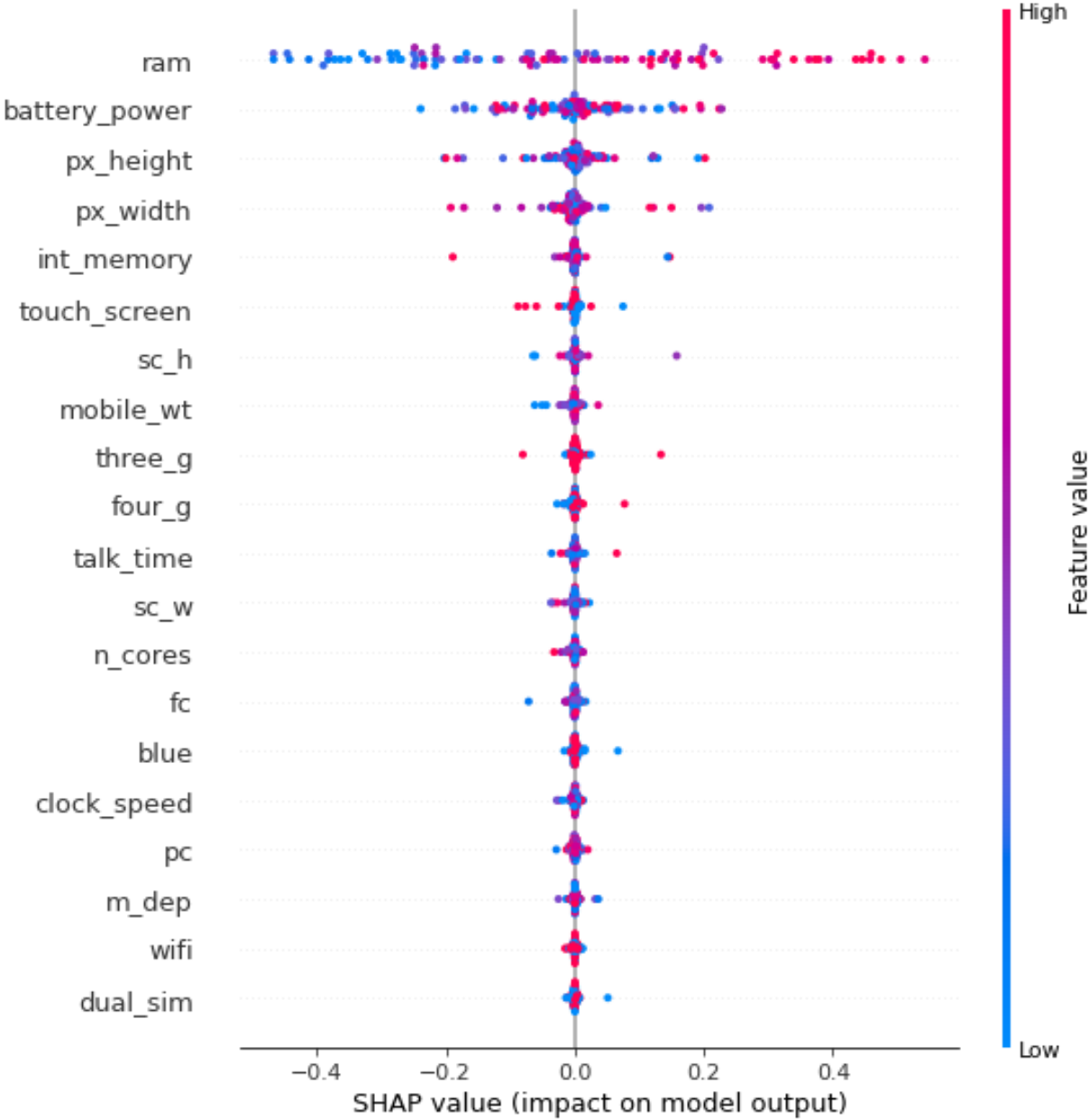

Figure 8: SHAP summary plot for 100 training samples using our proposed method on the MLP Classifier.

Form the results, it is observed that the proposed method performed well on both tree-based models and neural networks. First, we trained a random forest classifier on the mobile price classification data. The generated force plots as in Figure 1 and 3 are similar. Also, the summary plot generated using 100 random points on these ML model using proposed method as in Figure 4 is similar to the one generated by the tree explainer method as in Figure 2. Secondly, an MLP Classifier was trained on the mobile price classification data and the obtained force plots by the kernel explainer as in Figure 5 is similar to our proposed method as in Figure 7. The summary plot generated using 100 random points on this model using the proposed method as in Figure 8 is not only at par with the kernel explainer method in Figure 6 but also was computationally faster. The kernel explainer took 79.934 seconds of computation while the proposed method achieved in 1.52 seconds in a same computing device for same dataset.

**Regression data**

Regression analysis and its implementations for the proposed method was performed using a molded case circuit breaker (MCCB) electrical data of a three-phase electrical distribution system. Thirty features were considered including current, voltage, active power, etc of the system with its energy consumption of energy as the output label. This data is available on request.

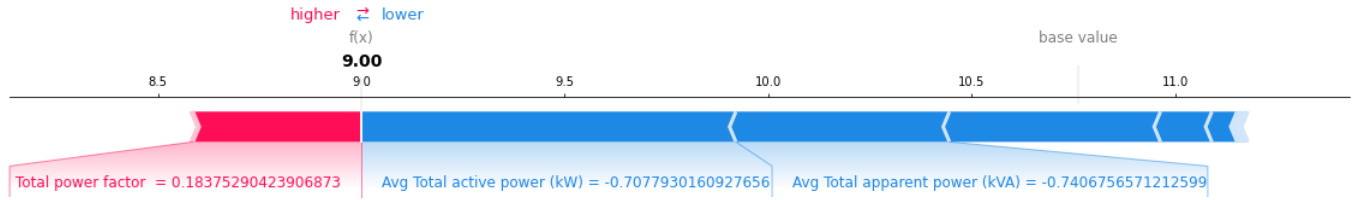

Figure 9: SHAP Force Plot for local interpretation using tree explainer on a Random Forest Regressor. It shows the feature avg total active power contributed the most in forcing the prediction in the negative direction followed by the feature avg total apparent power.

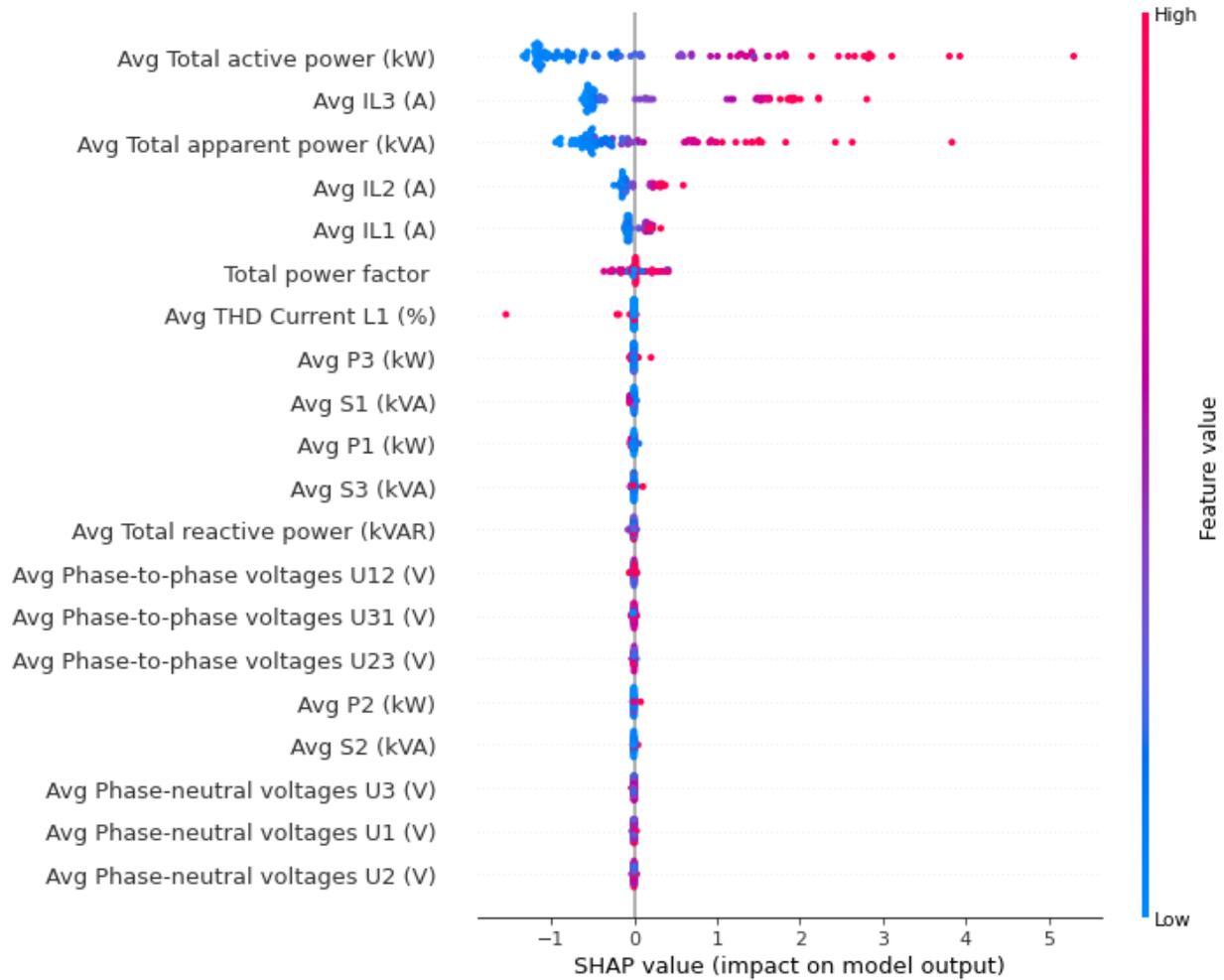

Figure 10: SHAP summary plot for 100 randomly selected samples using Tree Explainer on the Random Forest Regressor.

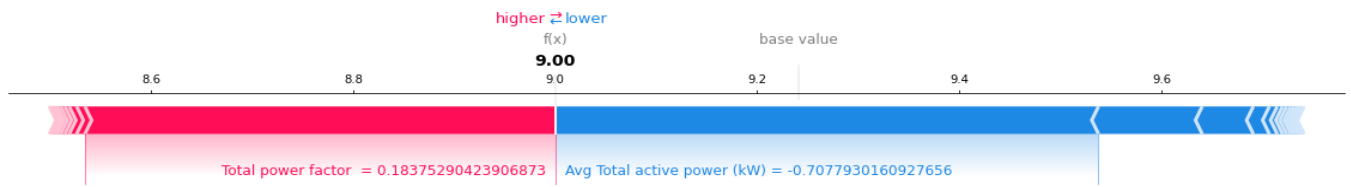

Figure 11: SHAP Force Plot for local interpretation using proposed method on a random forest regressor. We observe that the force plot drawn using proposed method is similar the tree explainer method.

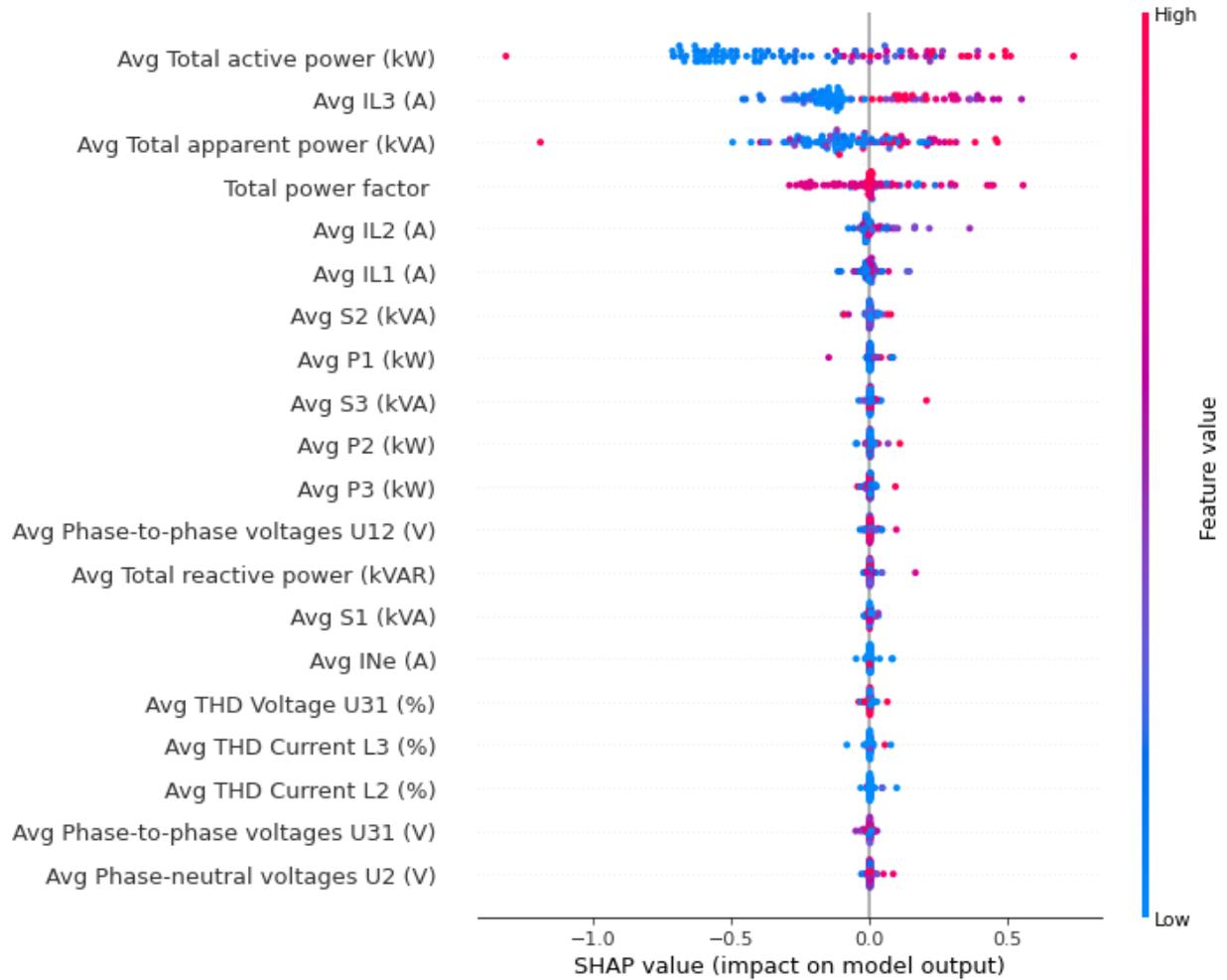

Figure 12: SHAP summary plot for 100 samples using the proposed method on random forest regressor shows the top 3 features are same as in the tree explainer.

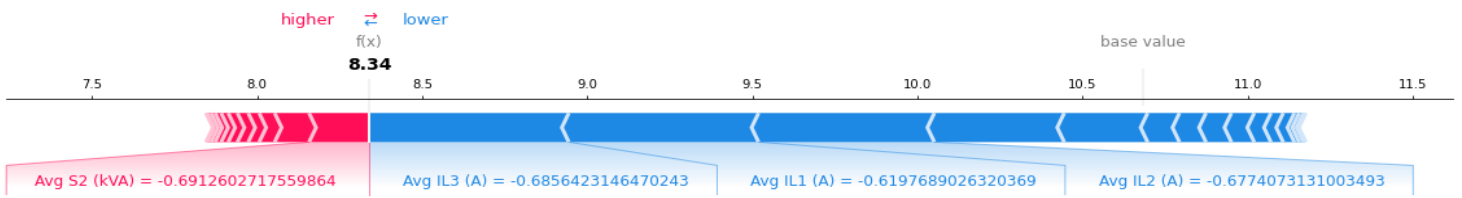

Figure 13: SHAP Force Plot for local interpretation using kernel explainer method on neural network. The computation time taken was 6.9 seconds.

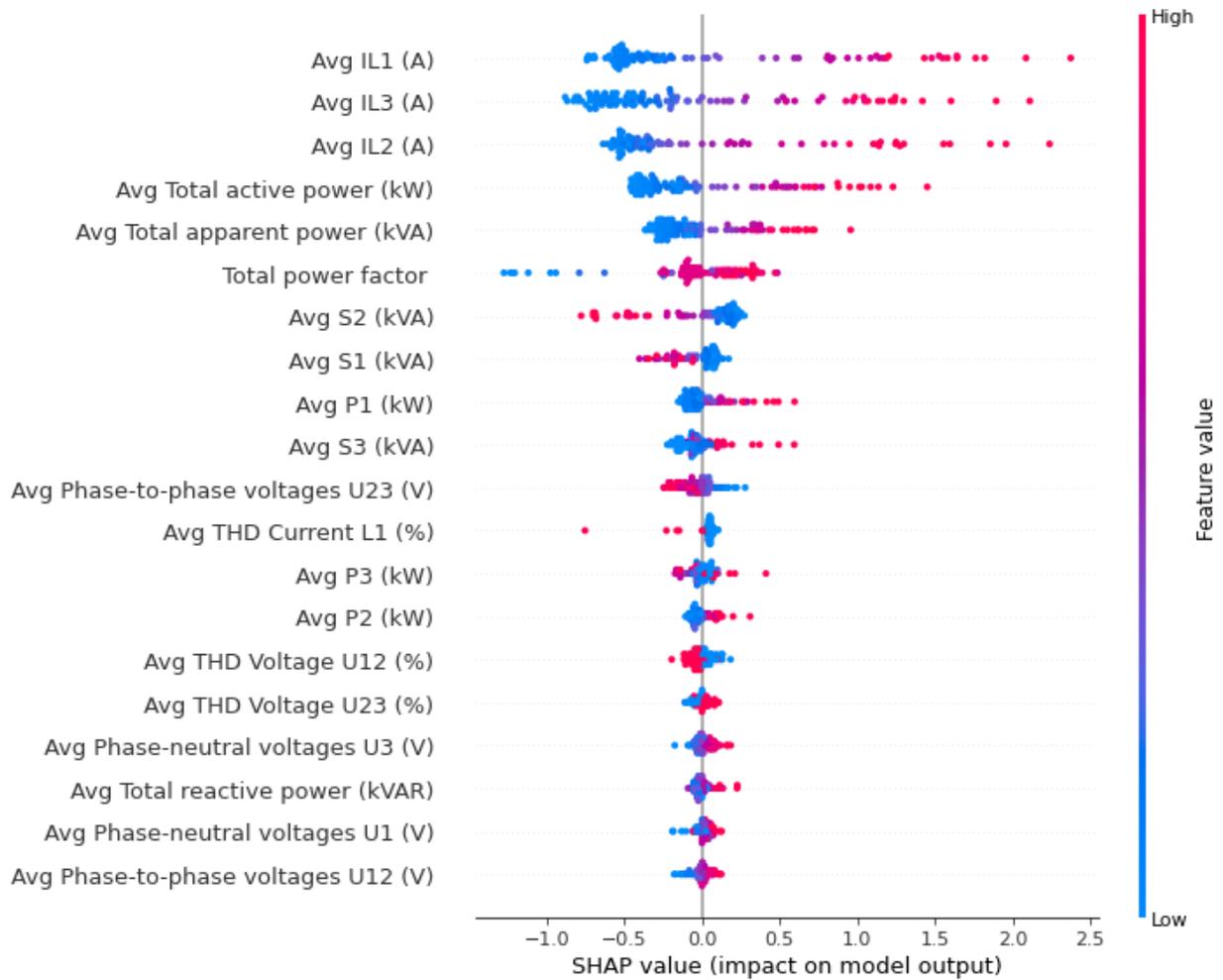

Figure 14: SHAP summary plot for 100 samples using kernel explainer method on the neural network. The computation time taken was 391.5 seconds.

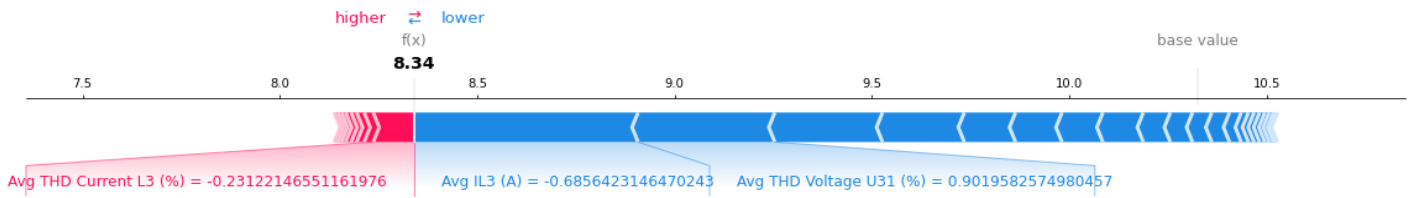

Figure 15: SHAP Force Plot for local interpretation using the proposed method on neural network. The computation time taken was 0.08 seconds.

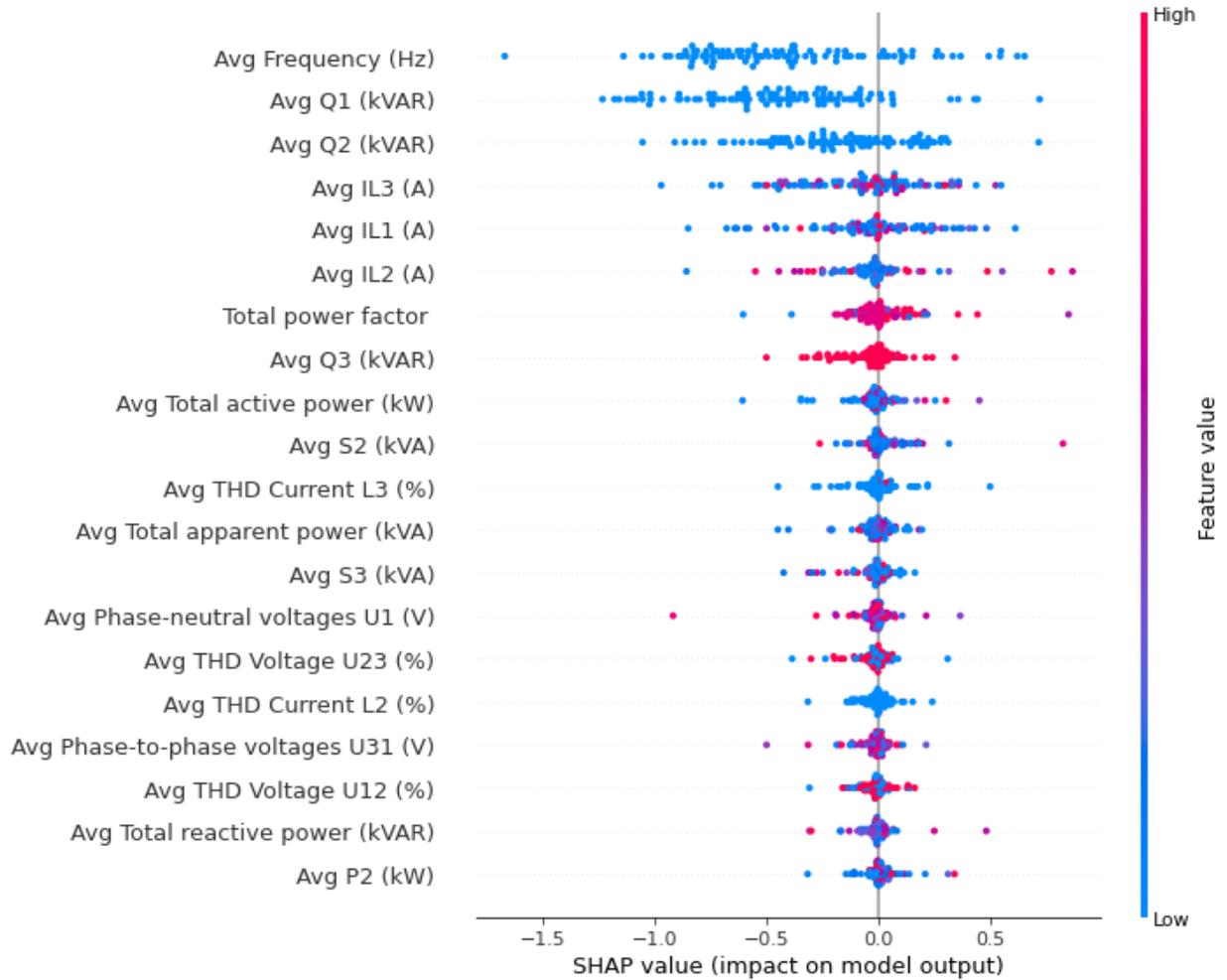

Figure 16: SHAP summary plot for 100 samples using the proposed method on the neural network. The computation time taken was 6.83 seconds.

## 4. Conclusion

In conclusion, the proposed LIMASE method was found to be capable of giving shapley explanation values on different models faster than the kernel explainer while being on par performance compared to SHAP tree explainer and the kernel explainer. The proposed method achieves it by training a locally faithful decision tree under the LIME paradigm and then uses the tree explainer on it to compute the Shapley explanation values. We also explained the proposed method ability to solve the submodular optimization problem for more efficient global interpretation and also its insight into regional interpretation.

## Authors' contributions

P. Sai Ram Aditya conceived the idea, performed the data analysis, and wrote the manuscript. Mayukha Pal conceptualized the project, contributed to idea generation, results analysis and discussion, guidance, and review of the manuscript.

## Declaration of Competing Interest

The authors declare that they have no known competing financial interest or personal relationships that could have appeared to influence the work reported in this paper. The data obtained for our analysis is from the available public domain database made for academic research purpose.

## Reference

1. Erico Tjoa, Cuntai Guan (2020). A Survey on Explainable Artificial Intelligence (XAI): Toward Medical XAI. IEEE Transactions on Neural Networks and Learning Systems. PP. 10.1109/TNNLS.2020.3027314. https://doi.org/10.1109/TNNLS.2020.3027314
2. Alejandro Barredo Arrieta, Natalia Díaz-Rodríguez, Javier Del Ser et al. (2019). Explainable Artificial Intelligence (XAI): Concepts, taxonomies, opportunities and challenges toward responsible AI, Information Fusion, Volume 58, 2020, Pages 82-115, ISSN 1566-2535. https://doi.org/10.1016/j.inffus.2019.12.012.
3. Sule Anjomshoae, Kary Främling, Amro Najjar (2019). Explanations of Black-Box Model Predictions by Contextual Importance and Utility. *ArXiv, abs/2006.00199*. https://doi.org/10.1007/978-3-030-30391-4_6
4. Maximilian Kohlbrenner, Alexander Bauer et al. (2020). Towards Best Practice in Explaining Neural Network Decisions with LRP. 1-7. 10.1109/IJCNN48605.2020.9206975. https://doi.org/10.48550/arXiv.1910.09840
5. Ioannis Kakogeorgiou, Konstantinos Karantzalos (2021). Evaluating explainable artificial intelligence methods for multi-label deep learning classification tasks in remote sensing, International Journal of Applied Earth Observation and Geoinformation, Volume 103, 2021, 102520, ISSN 0303-2434. https://doi.org/10.1016/j.jag.2021.102520
6. Valerio La Gatta, Vincenzo Moscato, Marco Postiglione, Giancarlo Sperlì (2021). CASTLE: Cluster-aided space transformation for local explanations, Expert Systems with Applications, Volume 179, 2021, 115045, ISSN 0957-4174. https://doi.org/10.1016/j.eswa.2021.115045.
7. Darius Afchar, Romain Hennequin, Vincent Guigue (2021). Towards Rigorous Interpretations: a Formalisation of Feature Attribution. *ArXiv, abs/2104.12437*. https://doi.org/10.48550/arXiv.2104.12437
8. Kacper Sokol, Alexander Hepburn, Raul Santos-Rodriguez, Peter Flach (2019). bLIMEy: Surrogate Prediction Explanations Beyond LIME. https://doi.org/10.48550/arXiv.1910.13016
9. Scott Lundberg and Su-In Lee (2017). A unified approach to interpreting model predictions. In proceedings of the 31st International Conference on Neural Information Processing Systems (NIPS'17). Curran Associates Inc., Red Hook, NY, USA, 4768-4777. https://doi.org/10.48550/arXiv.1705.07874
10. Jürgen Dieber, Sabrina Kirrane (2020). Why model why? Assessing the strengths and limitations of LIME. *ArXiv, abs/2012.00093*. https://doi.org/10.48550/arXiv.2012.00093

11. David Watson, Limor Gultchin, Ankur Taly, Luciano Floridi (2021). Local Explanations via Necessity and Sufficiency: Unifying Theory and Practice. https://doi.org/10.48550/arXiv.2103.14651
12. Elvio Amparore, Alan Perotti, Paolo Bajardi (2021). To trust or not to trust an explanation: using LEAF to evaluate local linear XAI methods. PeerJ Computer Science. 7. e479. 10.7717/peerj-cs.479. https://doi.org/10.7717/peerj-cs.479
13. Bence Mark Halpern, Finnian Kelly, Rob van Son, Anil Alexander (2020). Residual Networks for Resisting Noise: Analysis of an Embeddings-based Spoofing Countermeasure. *Odyssey*. http://dx.doi.org/10.21437/Odyssey.2020-46
14. Samuele Poppi, Marcella Cornia, Lorenzo Baraldi, Rita Cucchiara (2021). Revisiting The Evaluation of Class Activation Mapping for Explainability: A Novel Metric and Experimental Analysis. 2299-2304. 10.1109/CVPRW53098.2021.00260. https://doi.org/10.48550/arXiv.2104.10252
15. Marco Tulio Riberio, Sameer Singh, and Carlos Guestrin (2016). "Why Should I Trust You?": Explaining the Predictions of Any Classifier. In Proceedings of the 22$^{nd}$ ACM SIGKDD International Conference on Knowledge Discovery and data Mining (KDD'16). Association for Computing Machinery, New York, NY, USA, 1135-1144. https://doi.org/10.1145/2939672.2939778
16. Daniel Fryer, Inga Strumke, and Hien Nguyen (2021). Shapley Values for Feature Selection: The Good, the Bad, and the Axioms. *IEEE Access, 9*, 144352-144360. https://doi.org/10.48550/arXiv.2102.10936
17. Kjersti Aas, Martin Jullum, Anders Løland (2020). Explaining individual predictions when features are dependent: More accurate approximations to Shapley values, Artificial Intelligence, Volume 298, 2021, 103502, ISSN 0004-3702. https://doi.org/10.1016/j.artint.2021.103502
18. Alexandre Heuillet, Fabien Couthouis, Natalia Díaz-Rodríguez (2022). Collective eXplainable AI: Explaining Cooperative Strategies and Agent Contribution in Multiagent Reinforcement Learning With Shapley Values. IEEE Computational Intelligence Magazine. 17. 59 - 71. 10.1109/MCI.2021.3129959. https://doi.org/10.48550/arXiv.2110.01307
19. Scott Lundberg, Gabriel Erion, Hugh Chen *et al.* (2020). From local explanations to global understanding with explainable AI for trees. *Nat Mach Intell* **2,** 56–67. https://doi.org/10.1038/s42256-019-0138-9
20. Quan-shi Zhang Song-chun Zhu (2018). Visual interpretability for deep learning: a survey. *Frontiers Inf Technol Electronic Eng* **19,** 27–39. https://doi.org/10.1631/FITEE.1700808.
21. The Shapley Value explanations: https://towardsdatascience.com/the-shapley-value-for-ml-models-f1100bff78d1
22. Christoph Molnar (2022). Interpretable Machine Learning: A Guide for Making Black Box Models Explainable (2nd ed.). christophm.github.io/interpretable-ml-book/

23. Mobile price Classification data from Kaggle: https://www.kaggle.com/datasets/iabhishekofficial/mobile-price-classification